\documentclass[10pt,twocolumn,letterpaper]{article}

\usepackage{iccv}              
\usepackage[accsupp]{axessibility}  
\usepackage{graphicx}
\usepackage{booktabs}
\usepackage{multirow}
\usepackage{makecell}
\usepackage{xcolor}
\usepackage{amsmath}
\usepackage{colortbl}
\usepackage{diagbox}
\usepackage{soul}
\definecolor{colorfirst}{rgb}{.866,.945, 0.831} 
\definecolor{colorsecond}{rgb}{1, 0.98, 0.83} 
\definecolor{colorthird}{rgb}{0.76, 0.87, 0.92} 

\newcommand{\cellfirst}{\cellcolor{colorfirst}}
\newcommand{\cellsecond}{\cellcolor{colorsecond}}

\DeclareRobustCommand{\textfirst}[1]{\sethlcolor{colorfirst}\hl{#1}}
\DeclareRobustCommand{\secondtext}[1]{\sethlcolor{colorsecond}\hl{#1}}

\newcommand{\shortname}{Motion-2-to-3}
\newcommand{\twodnet}{2D Motion Diffusion model}

\newcommand{\mvnet}{Multi-view Diffusion model}

\newcommand{\twodnetsymbol}{\mathcal{D}_{2D}}

\newcommand{\mvnetsymbol}{\mathcal{D}_{mv}}

\newcommand{\supp}{the supplementary material}

\newcommand{\PAR}[1]{\vskip4pt \noindent{\bf #1~}}

\newcommand{\realnum}{\mathbb{R}}

\newcommand{\mtwo}{\mathcal{M}}

\def\Tabref#1{Table~\ref{#1}}

\def\Figref#1{Figure~\ref{#1}}

\def\Secref#1{Section~\ref{#1}}

\def\eqref#1{equation~\ref{#1}}

\definecolor{iccvblue}{rgb}{0.21,0.49,0.74}
\usepackage[pagebackref,breaklinks,colorlinks,allcolors=iccvblue]{hyperref}

\def\paperID{5677} 
\def\confName{ICCV}
\def\confYear{2025}

\title{Motion-2-to-3: Leveraging 2D Motion Data for 3D Motion Generations}
\author{
    Ruoxi Guo$^{1,2}$$^*$
    \quad Huaijin Pi$^{3}$$^*$
    \quad Zehong Shen$^{1}$
    \quad Qing Shuai$^{1}$
    \quad Zechen Hu$^{2}$
    \quad Zhumei Wang$^{2}$
    \\
    \quad Yajiao Dong$^{2}$
    \quad Ruizhen Hu$^{4}$
    \quad Taku Komura$^{3}$
    \quad Sida Peng$^{1}$
    \quad Xiaowei Zhou$^{1}$$^\dagger$
    \\[2mm]
    $^{1}$Zhejiang University
    \quad $^{2}$Deep Glint
    \quad $^{3}$The University of Hong Kong
    \quad $^{4}$Shenzhen University
}
\makeatletter
\newcommand\blfootnote[1]{%
  \begingroup
  \renewcommand\thefootnote{\relax}
  \footnotetext{%
    \let\hangfootparskip\relax       
    \setlength{\parindent}{0pt}
    #1
  }%
  \endgroup
}
\makeatother
\usepackage{footmisc}
\begin{document}
\twocolumn[\maketitle\vspace{-1.2em}\begin{center}
    \includegraphics[width=\linewidth]{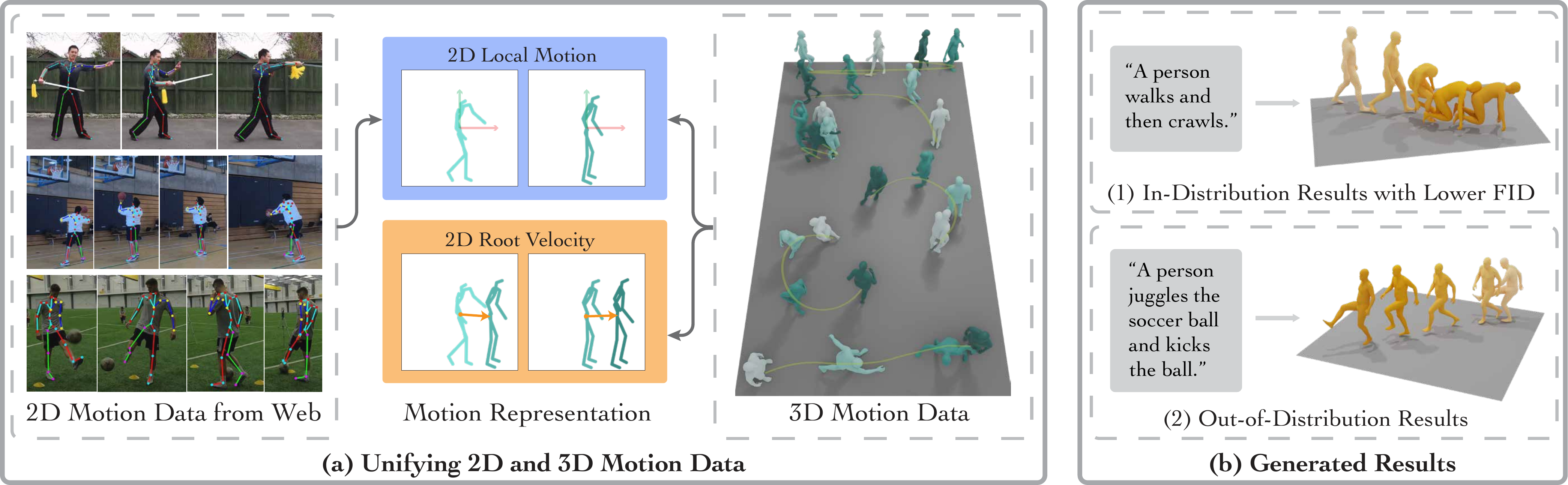}
\end{center}
\captionof{figure} {\textbf{Illustration of our key idea.}
    (a) Our approach leverages 2D motion data to improve 3D motion generation by unifying 2D and 3D motion data.
    (b) Our framework yields better FID and generates a broader range of motion types.}
\label{fig:teaser}\bigbreak]

\blfootnote{$\star$ Equal contribution to this work. \\$\dagger$ Corresponding author: Xiaowei Zhou.}
\begin{abstract}
Text-driven human motion synthesis has showcased its potential for revolutionizing motion design in the movie and game industry.
Existing methods often rely on 3D motion capture data, which requires special setups, resulting in high costs for data acquisition, ultimately limiting the diversity and scope of human motion. In contrast, 2D human videos offer a vast and accessible source of motion data, covering a wider range of styles and activities.

In this paper, we explore the use of 2D human motion extracted from videos as an alternative data source to improve text-driven 3D motion generation.
Our approach introduces a novel framework that disentangles local joint motion from global movements, enabling efficient learning of local motion priors from 2D data.
We first train a single-view 2D local motion generator on a large dataset of text-2D motion pairs.
Then we fine-tune the generator with 3D data, transforming it into a multi-view generator that predicts view-consistent local joint motion and root dynamics.
Evaluations on the well-acknowledged dataset and novel text prompts demonstrate that our method can efficiently utilize 2D data, supporting a wider range of realistic 3D human motion generation.
\end{abstract}    
\section{Introduction}
\label{sec:intro}
Text-driven human motion synthesis has drawn a great deal of attention in the 3D vision community~\cite{22cvpr_humanml3d}, with possibilities for applications across fields including movie production, gaming, and virtual reality. By providing an intuitive and simple interface for motion synthesis, Text-to-Motion (T2M) has the potential to streamline the character animation pipeline, reducing the need for extensive manual work.
It could also enhance immersive interactions in virtual environments by providing more dynamic and contextually relevant movements for avatars, bringing new responsive digital experiences.

Existing text-driven human motion generation techniques~\cite{22iclr_mdm,23cvpr_t2mgpt, 24cvpr_momask} almost exclusively rely on 3D motion datasets \cite{22cvpr_humanml3d}, which are primarily collected in marker-based high-quality motion capture systems~\cite{optitrack, vicon}.
Due to the need for specialized setups in controlled environments, 3D human motion datasets \cite{22cvpr_humanml3d} are limited in both size and diversity ~\cite{23iccv_maa}. 
The limited scope of the controlled collection condition limits the training set, failing to capture the true distribution of real-world movement. As a result, trained models struggle to scale and generalize, hindering their application in diverse environments.

In contrast, 2D human videos provide an affordable and widely accessible source of motion data.
They cover a wider range of motion styles and actions, reflecting diverse movements in natural settings, which can potentially augment the bias of 3D human motion capture data.  

In this paper, we focus on using 2D motion data extracted from 2D human videos to improve 3D motion generation. Using 2D data to assist 3D has been shown to be effective in static 3D object generation. 
For example, previous work \cite{23iccv_zero1to3,23arxiv_mvdream,22arxiv_dreamfusion} demonstrates that pre-training an image generative model on a large collection of 2D images before
fine-tuned on a smaller set of 3D data enabling a better recovery of photorealistic details compared to methods trained solely on limited 3D data.
We introduce a novel framework, called \shortname{}, which utilizes 2D data to enhance 3D human motion generation. 
Our key insight is to degrade the 3D motion generation task to multi-view 2d motion generation, by introducing a multi-view 2D motion generator to maintain view consistency. We first collect a large-scale dataset of text-video pairs and extract 2D motion sequences to train a single-view 2D local motion generator. To generate 3D human motion, we fine-tune the single-view 2D motion generator with 3D data, adapting it into a multi-view 2D motion generator. To be more specific, we enhance each transformer layer in the 2D motion generator by adding a view attention layer, enabling simultaneous multi-view generation. 

\begin{figure}[tbp]
    \centering
    \includegraphics[width=0.95\linewidth]{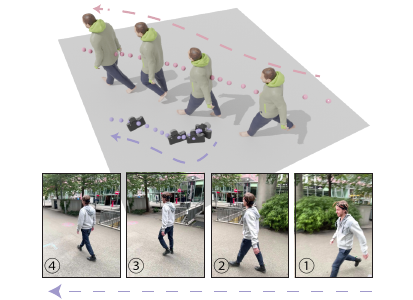}
    
    \caption{\textbf{Challenge of 2D motion from the real world.}
    In the real-world videos \cite{23iccv_emdb}, both the camera and humans move in 3D space, resulting in 2D motion that combines both movements.}
    \label{fig:challenge}
\end{figure}

However, it is not straightforward to extend the multi-view training strategies used in 3D object generation~\cite{23arxiv_mvdream,23iccv_zero1to3} to 3D human motion generation.
In general, 2D human motion data cannot accurately reflect real-world 3D human motion, as 2D motion typically entangles camera movement and 3D human motion, as illustrated in 
\Figref{fig:challenge}

Therefore, we make a major innovation to disentangle local human motion from global human movement, enabling us to accurately learn local motion priors from 2D data.
Specifically, we reformulate 2D human motion as 2D local motion and root velocity sequences.
This approach effectively circumvents the problem of inaccurate global movement through the separation of 2D motion components, making efficient use of abundant 2D data.

With disentangled local pose and root velocity sequences, we add a root velocity head to predict the 2D root movement specific to each view.
Using 3D data, we could create synthetic multi-view 2D motion sequences by projecting 3D local motion and root velocities into different views.
These multi-view sequences serve as training data, enabling the generator to produce coherent global movement while maintaining view consistency.

In summary, our contributions are threefold: 1. We introduce a novel framework, \shortname{}, that leverages 2D motion data from human videos to enhance 3D human motion generation. 2. We propose a method to disentangle local human motion from global human movement, enabling accurate learning of local motion priors from abundant 2D data. 3. We demonstrate the effectiveness of our approach through extensive experiments on the HumanML3D dataset \cite{22cvpr_humanml3d}, providing both quantitative and qualitative results.
Additionally, we use varied out-of-distribution (OOD) text prompts to validate our framework’s ability to generate a broader range of motion types. Experimental results show that it significantly improves the quality and diversity of 3D motion generation compared to methods relying solely on limited 3D data.

\section{Related work}
\label{sec:relatedwork}
\subsection{Text-driven motion generation}
Recently, there has been a significant rise in research focusing on text-driven motion generation \cite{193dv_l2p, 22eccv_temos, 22eccv_motionclip}.
This task takes natural language descriptions as input and synthesizes human motion that accurately reflects the provided instructions.
The first benchmark dataset for this task, KIT-ML \cite{16bigdata_kit}, laid the groundwork for subsequent studies.
Following this, BABEL \cite{21cvpr_babel} provides per-frame labels for the AMASS dataset \cite{19iccv_amass}.
In addition, HumanML3D \cite{22cvpr_humanml3d} annotates the dataset with sequence-level descriptions.
Moreover, Motion-X \cite{24nips_motionx} contributes a comprehensive 3D whole-body human motion dataset.

Several approaches have been proposed to tackle this task.
Early attempts \cite{193dv_l2p, 21iccv_scatd, 22eccv_motionclip} aim to learn a shared latent space for both text and motion.
For instance, TEMOS \cite{22eccv_temos}, utilizes a transformer-based VAE \cite{21iccv_actor} to generate diverse motion outputs.
Additionally, methods such as TM2T \cite{22eccv_tm2t} and T2M-GPT \cite{23cvpr_t2mgpt} achieve enhanced performance through the use of discrete representations via VQ-VAE \cite{17nips_vqvae, 19nips_vqvae2}.
Other work \cite{22iclr_mdm, 23aaai_flame, 22arxiv_motiondiffuse, 23cvpr_mofusion} like MDM \cite{22iclr_mdm} have successfully applied diffusion models \cite{20nips_ddpm} in this direction.
Further explorations of latent diffusion models \cite{22cvpr_stablediffusion} are seen in \cite{23cvpr_belfusion, 23iccv_priority, 23cvpr_mld}.
Building upon MDM \cite{22iclr_mdm}, methods like \cite{23iccv_gmd, 24iclr_omnicontrol, 24iclr_priormdm} introduce sparse control.
Other existing works \cite{23iccv_sinc, 23nips_motiongpt_tencent, 24aaai_motiongpt} leverage Large Language Models (LLM) \cite{23arxiv_llama, 23arxiv_gpt4, 20nips_llmfewshot, 22nips_llmzeroshot, 20jmlr_t5} in the motion domain to support various motion tasks \cite{23nips_motiongpt_tencent}.
More recent work \cite{24cvpr_momask} further explores residual VQ \cite{22cvpr_resvq} and generative masked modeling \cite{22cvpr_maskgit,23cvpr_mage}.
MotionMamba \cite{24eccv_motionmamba} successfully applies Mamba \cite{23arxiv_mamba} in the motion domain. MotionStreamer\cite{xiao2025motionstreamer} incorporates a continuous causal latent space into a probabilistic autoregressive model. 
Additionally, \cite{22nips_humanise, 23iccv_hghoi, 24eccv_tesmo, 24cvpr_scenetext} generate motion with scene information.
Some work \cite{22eccv_motionclip, 22sig_avatarclip} also explores open-vocabulary text-to-motion generation.
For example, MotionCLIP \cite{22eccv_motionclip} and AvatarCLIP \cite{22sig_avatarclip} rely on the CLIP \cite{21icml_clip} latent space.
MAA \cite{23iccv_maa} uses 3D poses estimated from large-scale image-text datasets to pretrain the model.
OOHMG \cite{23cvpr_beingcomes} proposes a method in a zero-shot learning manner that does not require paired text-motion training data by reconstructing motion from keyframes.
PPG \cite{24eccv_ppg} uses ChatGPT \cite{22arxiv_chatgpt} to help keyframe pose generation and then generates motion from these keyframes.
OMG \cite{24cvpr_omg} employs existing 3D motion data without text \cite{19iccv_amass} to pretrain a model and then finetune it on the HumanML3D \cite{22cvpr_humanml3d} dataset using \cite{23iccv_controlnet}.

MAS \cite{24cvpr_mas} is the first work to leverage 2D motion data from videos to generate 3D motion.
A 2D motion diffusion model \cite{22iclr_mdm} is trained and MAS uses it to generate multi-view 2D motion independently.
During the diffusion process, they propose a consistency block to enforce the view consistency, by converting the 2D motion to 3D motion using triangulation \cite{97_triangulation} and projecting back.
TENDER \cite{24arxiv_tender} further collects a larger dataset to train a 2D motion model.
However, MAS \cite{24cvpr_mas} only considers some specific motion types without text control.
They \cite{24cvpr_mas,24arxiv_tender} only generate some local 2D motion, and the generated 3D motion may have artifacts due to the inconsistency among multi-view results.
In contrast, we train a text-driven multi-view 2D motion generator and enable global movements, with consistent multi-view results.

\subsection{3D generation}
DreamFusion \cite{22arxiv_dreamfusion} and SJC \cite{23cvpr_sjc} apply 2D diffusion priors \cite{22cvpr_stablediffusion} to generate a 3D object.
The key technique is called score distillation sampling (SDS), where the diffusion priors supervise the optimization of a 3D representation \cite{20eccv_nerf}.
Following works try to improve the performance by introducing better 3D representations \cite{23cvpr_magic3d,243dv_textmesh,23iccv_makeit3d,23iccv_fantasia3d,23arxiv_dreamgaussian} and loss designs \cite{24nips_prolificdreamer,23arxiv_nsfd,23arxiv_csd}.
Although these methods could generate photo-realistic results, they are known to suffer from multi-view consistency issues.
To overcome this, Zero1to3 \cite{23iccv_zero1to3} proposes to finetune the stable diffusion models \cite{22cvpr_stablediffusion} on 3D object datasets \cite{23cvpr_objaverse} to generate a novel view of an input image based on a relative camera pose.
\cite{23arxiv_mvdream,24nips_one2345} directly generate multi-view images from a single view input.
\cite{24cvpr_eschernet, 24arxiv_cat3d} generate multi-view images from a flexible number of reference views.
After training on 3D object datasets \cite{23cvpr_objaverse}, these methods could generate multi-view consistent 3D objects.
\cite{23arxiv_mav3d, 24cvpr_ayg, 24arxiv_sv4d} further explores dynamic 3D object generation by introducing video diffusion models \cite{23cvpr_ayl}.
However, they only demonstrate the ability to generate 3D objects with tiny movements.

Different from these methods, which only consider static objects or small movements, we focus on generating 3D human motion. 
We decouple the global movement and local pose changes and share the same insight as \cite{23arxiv_mvdream,24nips_one2345} to employ 3D data for multi-view consistent results.

\begin{figure*}[tbp]
    \centering
    \includegraphics[width=1.\linewidth]{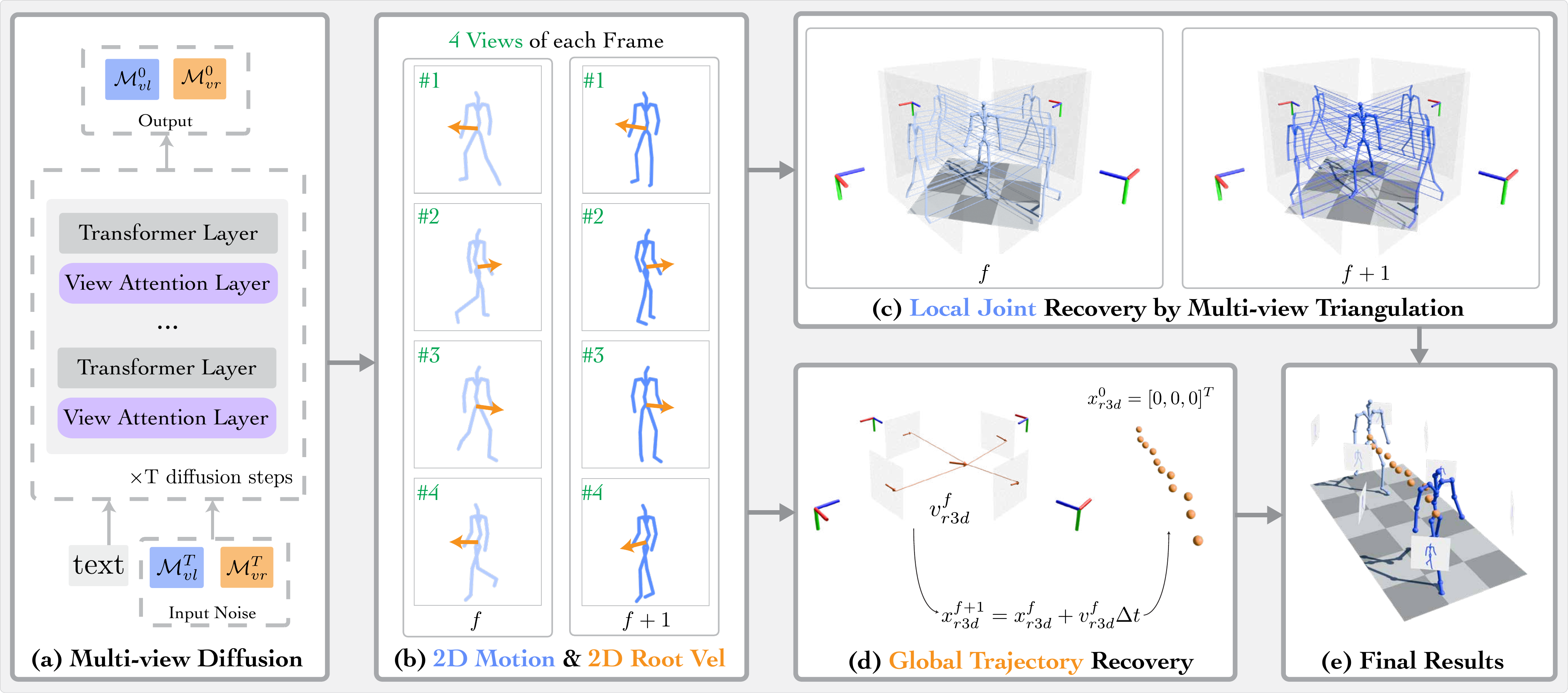}
    
    \caption{\textbf{Our Pipeline.}
    We design a \mvnet{} (a) to generate multi-view results (for simplicity, camera embedding is omitted in the figure).
    During inference, the \mvnet{} predicts 2D local motion and root velocity (b).
    Then, we use triangulation \cite{97_triangulation} to recover 3D local joint positions (c) and accumulate root velocity to obtain 3D global trajectory (d), resulting in the final 3D motion (e).
    }
    \label{fig:pipeline}
\end{figure*}
\section{Method}
\label{sec:method}
We propose a novel pipeline called \shortname{} to generate 3D motion from text input by leveraging multi-view 2D motion representations, as illustrated in \Figref{fig:pipeline}.
First, we describe how to train a single-view 2D motion generator, \twodnet{}, on disentangled 2D local motion extracted from video data to learn a local motion prior (\Secref{sec:2dmotion}).
Then, building on \twodnet{}, we introduce \mvnet{} to extend this model to generate multi-view consistent 2D motion, incorporating both view consistency and root movement (\Secref{sec:mvnet}).
Finally, we outline the inference process to recover 3D motion from multi-view 2D motion (\Secref{sec:inference}).

\subsection{2D motion generation}
\label{sec:2dmotion}
Here we focus on training a 2D human motion model using video data. 
In a video, each joint's motion is a blend of three components: camera movement, the global movement of the human body, and the local motion of each joint relative to the root.
The local motion captures each joint’s individual movement, while the root motion includes both the human’s global movement and camera motion.

Given a video of $N$ frames, we extract $J$ joints' 2D positions of each frame which compose 2D motion $\mtwo \in \realnum^{N \times J \times 2}$.
We decompose the 2D motion to root position and local motion \(\mtwo_l \in \realnum^{N \times (J-1) \times 2}\), where we obtain it by subtracting the root position from each joint’s position, effectively removing the influence of global movement and camera motion from each joint.
This decomposition enables us to focus solely on learning local motion components.

To train a 2D motion generator, we first extract 2D motion sequences using an off-the-shelf 2D pose estimator and then convert these sequences to local 2D motion.
Given the variation in video widths and heights, we normalize the 2D poses by calculating bounding boxes based on the extracted poses \cite{24cvpr_wham}.
Text annotations for the videos are mostly provided by the dataset.
Following MDM \cite{22iclr_mdm}, we employ a transformer-based \cite{17nips_attention} diffusion model \cite{20nips_ddpm} to train a \twodnet{} $\twodnetsymbol{}$.
The model input includes text embeddings \(\mathcal{T} \in \realnum^{77 \times 768}\) extracted by the CLIP \cite{21icml_clip} model, along with random noise.
The output is a sequence of 2D local motion, \(\mtwo_l \in \realnum^{N \times (J-1) \times 2}\), which represents joint movements disentangled from global and camera-induced motion.

\subsection{\mvnet{}}
\label{sec:mvnet}
Generating each view’s 2D motion independently using \twodnet{} \cite{24cvpr_mas} fails to ensure view consistency, leading to artifacts in the 3D motion.
To address these issues, we design a \mvnet{} that leverages 3D motion data to enforce view consistency across generated 2D motion.
Additionally, we incorporate a root velocity head to predict the 2D root movement for each view, enabling realistic global motion and coherent multi-view results.

To train the \mvnet{}, we synthesize multi-view 2D motion sequences from 3D motion data by projecting 3D motion into different camera views.
In the first frame, we initialize V virtual cameras randomly positioned around the character, each centered on the human root joint.
For subsequent frames, these cameras move along with the character’s root, ensuring that the root remains at the center of each view and maintaining consistent tracking across frames.
We project the 3D joint motion and root velocities into each camera view, yielding 2D local motion and root velocities specific to each perspective.
This setup allows the 2D motion from each view to capture joint-specific local movements, while the projected root velocity represents the global character movement. 
As a result, the multi-view data consistently reflects both local and global motion components.

Given the synthesized multi-view sequences, we train the \mvnet{} to generate consistent multi-view 2D motion sequences.
As illustrated in \Figref{fig:pipeline}(a), \mvnet{} extends \twodnet{} $\twodnetsymbol{}$ by adding multi-view attention layers and an additional root velocity head.
These modifications enable the model to capture view-specific information, maintain consistency across views, and incorporate realistic global movement.
Alongside text embeddings \(\mathcal{T} \in \realnum^{77 \times 768}\), the model input includes camera embeddings and noisy multi-view 2D motion at each diffusion step.
The camera embeddings \(\mathcal{C}_{rel} \in \realnum^{V \times 4}\) represent the relative poses of each view with respect to a randomly chosen first view, with \(V\) denoting the number of views.
While the first view is randomly selected, the relative arrangement of the other views remains fixed, capturing essential view-specific information.
The noisy multi-view 2D motion input comprises multi-view 2D local motion \(\mtwo^t_{vl} \in \realnum^{N \times V \times (J-1) \times 2}\) and multi-view 2D root velocity \(\mtwo^t_{vr} \in \realnum^{N \times V \times 1 \times 2}\), representing joint-specific and root movement information for each view.
The camera pose embeddings are projected into the latent space and added with the corresponding view’s motion at each diffusion step \(t\)
(for details, refer to \supp{}).
In each block, the multi-view attention layers operate across views to ensure consistency, while the transformer layers inherited from \twodnet{} focus on the temporal dimension.
The root velocity head predicts the 2D root movement for each view, enabling realistic global movements.
The model outputs the clean multi-view 2D local motion $\mtwo^0_{vl}$ and root velocity $\mtwo^0_{vr}$, as illustrated in \Figref{fig:pipeline} (b).
Following \cite{23iccv_controlnet}, we freeze the original layers from \(\twodnetsymbol{}\) and train only the newly added multi-view attention layers and root velocity head, focusing learning on enforcing view consistency and global movement.

\subsection{Inference}
\label{sec:inference}
Given the input text, we begin by initializing multiple cameras and using \mvnet{} $\mvnetsymbol{}$ to generate multi-view 2D motion sequences, as shown in \Figref{fig:pipeline} (b).
Following MAS \cite{24cvpr_mas}, we then apply triangulation \cite{97_triangulation} to convert these multi-view 2D motion into 3D local motion as illustrated in \Figref{fig:pipeline} (c).
As depicted in \Figref{fig:pipeline} (d), our decoupled representation allows us to compute 3D root velocity during triangulation.
We could accumulate the root velocity over time to get the root trajectory via
$x_{r 3 d}^{f+1}=x_{r 3 d}^f+v_{r 3 d}^f \Delta t$,
where
$x_{r 3 d}^{f}$ and $v_{r 3 d}^f$ denote the 3D root position and root velocity at $f$-th frame, and
$\Delta t$ is the time interval.
By combining the accumulated 3D root velocity with the 3D local poses, we generate a 3D motion sequence that includes global movement as shown in \Figref{fig:pipeline} (e).
Finally, to animate a rigged character, we follow prior methods \cite{22iclr_mdm,24cvpr_mas} and use SMPLify \cite{16eccv_smplify} to fit the SMPL \cite{15tog_smpl, 19cvpr_smplx} pose parameters.
\section{Experiment}
\label{sec:exp}
\subsection{Implementation details}
We use $8$ transformer decoder layers \cite{17nips_attention} to construct the \twodnet{}.
Each layer has $4$ heads and $512$ hidden units, and the feed-forward layer has $1024$ hidden units.
We first train \twodnet{} on synthetic 2D motion with a learning rate of $0.0001$ and a batch size of $128$ for $100$ epochs.
Then we train it on all 2D motion data (102,389 sequences) with a learning rate of $0.00001$ and a batch size of $128$ for $100$ epochs.
Next, we add an extra multi-view attention layer in each transformer decoder layer and an additional MLP layer for root velocity on \twodnet{} to build the \mvnet{} with $V=4$.
The blocks from \twodnet{} are frozen and \mvnet{} is finetuned for $100$ epochs with a learning rate of $0.0001$ and a batch size of $32$.
We use the Adam optimizer \cite{14arxiv_adam} for the training.

\subsection{Datasets}
\PAR{2D data collection.}
We use the human videos from open-source datasets EgoExo4D \cite{24cvpr_egoexo4d} and Motion-X++ \cite{24nips_motionx} to avoid potential privacy concerns.
We employ a commercially used human detection model and a pose estimation model from a company to extract 2D poses in SMPL \cite{15tog_smpl} skeleton from the videos.
Then, we use the human tracking algorithm from EasyMocap \cite{21_easymocap} to track the human.
The 2D poses are further smoothed by SmoothNet \cite{22eccv_smoothnet} to obtain the final 2D motion.
We also filter out the extremely short motion and remove joints with low detection confidence.
We normalize 2D motion with the bounding box of the human following \cite{24cvpr_wham}.
EgoExo \cite{24cvpr_egoexo4d} and Motion-X++ \cite{24nips_motionx} already provide text annotations for the videos. 
Due to different language styles across datasets, we apply ChatGPT-3.5 \cite{22arxiv_chatgpt} for text augmentation to harmonize the style.
In total, we use $97.63$ hours of human video.
For the detailed statistics of 2D data, please refer to \supp{}.

\PAR{3D dataset.}
The HumanML3D dataset~\cite{22cvpr_humanml3d} is widely used in previous motion generation tasks.
It collects 14,616 motion sequences from AMASS~\cite{19iccv_amass} and annotates 44,970 sequence-level textual descriptions.
The total duration is $28.59$ hours.
We use the dataset both for training and for evaluation. 
In the \twodnet{} training process, we randomly sample 2D motion using virtual cameras surrounding the character. 
In the \mvnet{}, we extract multi-view 2D motion of the 3D motion in the dataset.
For more details about synthetic cameras, please refer to \supp{}.

\subsection{Metrics}
On the HumanML3D \cite{22cvpr_humanml3d} dataset, we follow previous methods \cite{22iclr_mdm,22cvpr_humanml3d} using the following metrics:
(1) Motion-retrieval precision (R-Precision) calculates the text and motion matching accuracy among 32 sequences.
Top-3 accuracy of motion-to-text retrieval is reported.
(2) Fréchet Inception Distance (FID) measures the feature distributions between the generated and real motion. 
(3) Multimodal Distance (MM Dist) calculates average distances between each text feature and the generated motion features.
(4) Diversity: we compute the average Euclidean distance between motion features from 300 randomly sampled motion pairs.
Similar to \cite{22cvpr_humanml3d,22iclr_mdm}, we transform the generated motion into the 263-dimensional pose representation vector \cite{22cvpr_humanml3d} to calculate the above metrics.

We also invite 56 participants from different institutes to evaluate the generated motion.
Each participant was presented with motions generated from 15 novel text prompts and was asked to select the best and second-best motion from each group.
Out of the 56 responses, 49 were deemed valid, with incomplete questionnaires excluded.
This portion of the test data is denoted as \textit{novel text}.

\subsection{Main results}
\begin{table}[tbp]
    \centering
    \resizebox{0.8\linewidth}{!}{%
    \begin{tabular}{lccccc}
    \toprule
    Methods & R-Precision $\uparrow$&
    FID$\downarrow$ &
    MM Dist$\downarrow$&
    Diversity$\rightarrow$\\
    \midrule
    Real                                                    & $0.797$ & $0.002$ & $2.974$ & $9.503$ \\
    \midrule
    MDM \cite{22iclr_mdm}                                   & $0.611$ & $0.544$ & $5.566$ & \cellfirst $\textbf{9.559}$ \\
    MLD \cite{23cvpr_mld}                                   & \cellsecond $0.772$ & $0.473$ & \cellfirst $\textbf{3.196}$ & $9.724$ \\
    MAA \cite{23iccv_maa}                                   & $0.675$ & $0.774$ & $-$     & $8.230$ \\
    OMG \cite{24cvpr_omg}                                   & \cellfirst $\textbf{0.784}$ & \cellsecond $0.381$  & $-$     & \cellsecond $9.657$ \\ 
    \midrule
    Ours                                                    & $0.697$ & \cellfirst $\textbf{0.321}$  & \cellsecond $3.579$ & $9.286$ \\
    \bottomrule
    \end{tabular}
    }
    \caption{\textbf{Main results of text-conditional motion synthesis on HumanML3D~\cite{22cvpr_humanml3d} dataset.}
    These metrics are evaluated by the motion encoder from \cite{22cvpr_humanml3d}. 
    The right arrow $\rightarrow$ means the closer to real motion the better.
    The dash $-$ denotes the results are unavailable as they do not release the code.
    The best and second-best results are highlighted \textfirst{\textbf{green}} and \secondtext{yellow}, respectively.
    }
    \label{tab:comp:humanml3d}
\end{table}
\begin{table}[tbp]
    \centering
    \resizebox{0.8\linewidth}{!}{
        \begin{tabular}{cccc}
            \toprule
            Metrics & Ours & MDM \cite{22iclr_mdm} & MLD \\  
            \midrule
            Best Motion Rate $\uparrow$ & \cellfirst $ \textbf{51.43\%} $ & $ 24.08\% $ & \cellsecond$ 24.49\%$\\
            Top-2 Motion Rate $\uparrow$ & \cellfirst $ \textbf{84.49\%} $ & \cellsecond $ 60.68\% $ & $ 54.83\% $ \\
            \bottomrule
        \end{tabular}
    }
    \caption{
    \textbf{User study on \textit{novel text} prompts.}
    Best Motion Rate and Top-2 Motion Rate represent the proportions of being selected as the best motion and as one of the top two motions.
    If selected randomly, the expected rate would be $33\%$.
    }
    \label{tab:comp:novel}
    \vspace{-2ex}
\end{table}
We compare our approach with various state-of-the-art methods on the HumanML3D dataset \cite{22cvpr_humanml3d}.
We select most representative diffusion-based methods trained with 3D representation~\cite{22cvpr_humanml3d}, including MDM~\cite{22iclr_mdm}, MLD~\cite{23cvpr_mld}, MAA~\cite{23iccv_maa}, and OMG~\cite{24cvpr_omg}.
MDM~\cite{22iclr_mdm} is the first work that employs diffusion models \cite{20nips_ddpm} for motion generation and we use a similar architecture to them.
MLD~\cite{23cvpr_mld} employs latent diffusion \cite{22cvpr_stablediffusion} to generate human motion.
MAA~\cite{23iccv_maa} uses 3D poses estimated from large-scale image-text datasets to pretrain the model.
OMG~\cite{24cvpr_omg} employs existing 3D motion data without text~\cite{19iccv_amass} to pretrain a model and then finetune it on the HumanML3D~\cite{22cvpr_humanml3d} dataset.
We also evaluate the performance of the novel text prompts and compare them with open-sourced methods MDM~\cite{22iclr_mdm} and MLD~\cite{23cvpr_mld}.

The quantitative results on the HumanML3D \cite{22cvpr_humanml3d} are presented in \Tabref{tab:comp:humanml3d} and the novel text results are shown in \Tabref{tab:comp:novel}.
Our approach obtains better FID than baselines and comparable performance in other metrics.
Unlike the baseline methods \cite{22iclr_mdm, 23cvpr_mld,24cvpr_omg}, our model is not directly optimized on the 3D representation used for evaluation, which may account for the slightly lower scores on three of the four metrics.
This difference stems from the metrics’ reliance on the 3D feature space: FID measures distribution distance, while the other three are element-wise, favoring models optimized in 3D.
We also observe that our method produces lower motion diversity than some baselines, possibly because training with additional text-motion pairs strengthens the mapping between text and motion, thus constraining diversity. 
A similar effect has been noted in MAA \cite{23iccv_maa}, where using text-pose pairs also reduced diversity.
Besides, our method achieves the best user study result, demonstrating its ability to learn a wider variety of motion from 2D data, which enables it to perform better on \textit{novel text} inputs.

    

    

\begin{figure*}[tbp]
    \centering
    \includegraphics[width=1.0\linewidth]{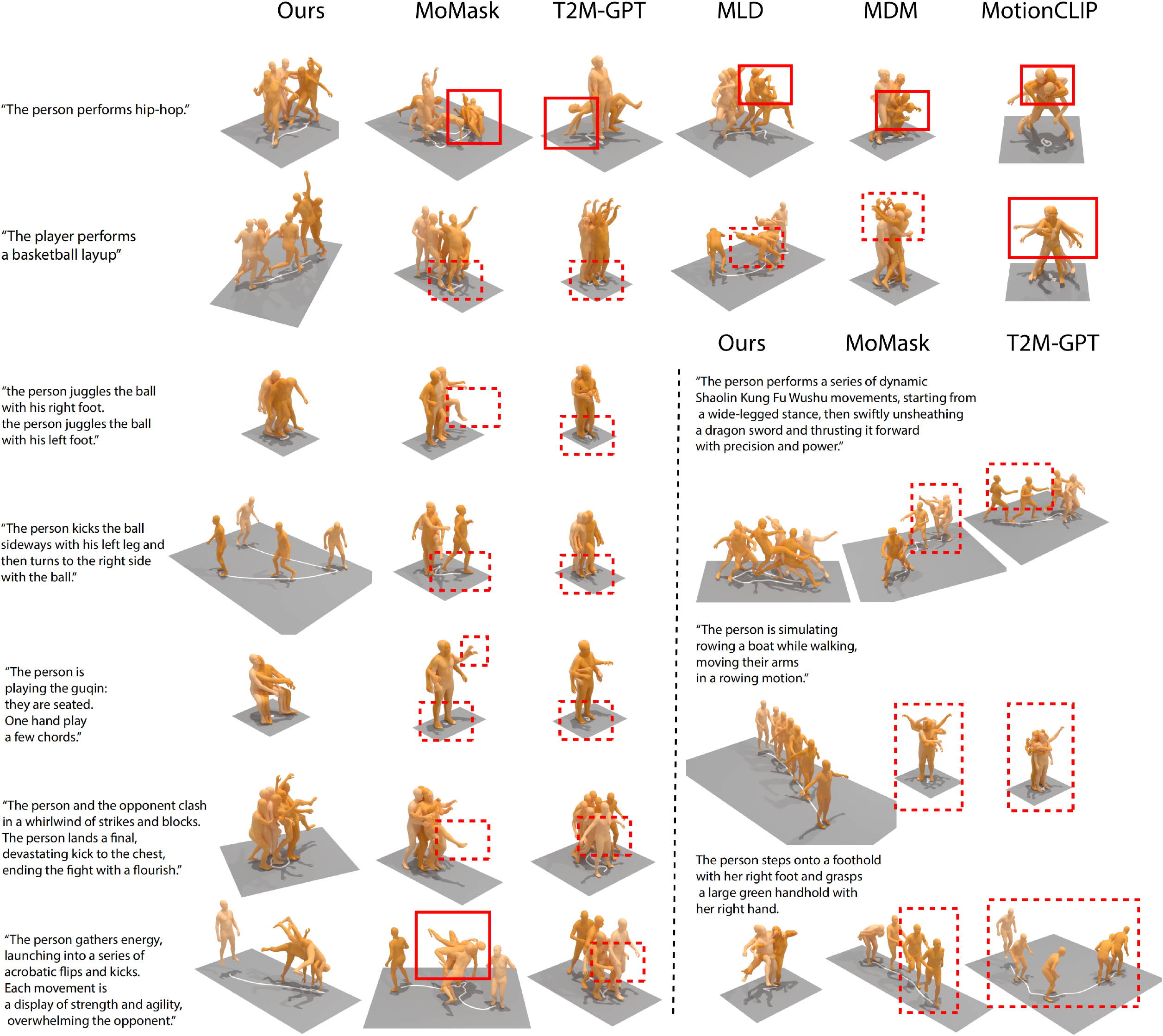}
    
    \caption{\textbf{Qualitative comparison.}
    The first two lines are motion out of the HumanML3D \cite{22cvpr_humanml3d} dataset. We compare our quality with all the baseline datasets. Baseline methods produce incorrect types of motion, while ours are more consistent with the text descriptions. 
    The following rows are novel text prompt results. We compare our quality with the SOTA methods MoMask and T2M-GPT\cite{24cvpr_momask, 23cvpr_t2mgpt}. When encountered with novel text, other methods failed to produce correct types of motion, while ours are more consistent with the text descriptions. 
    The unnatural poses are highlighted in the red boxes.
    The semantics misalignment is highlighted in the dashed boxes. 
    }
    \label{fig:comp}
\end{figure*}
The qualitative results are shown in \Figref{fig:comp}.
We observe that baseline methods often generate incorrect motion types that do not align well with the text prompts.
In contrast, our method generates motion that is consistent with the text descriptions.
This is because our method leverages the 2D motion data, which captures a larger variety of human motion than the 3D data. 

\begin{figure*}[tbp]
    \centering
    \includegraphics[width=0.85\linewidth]{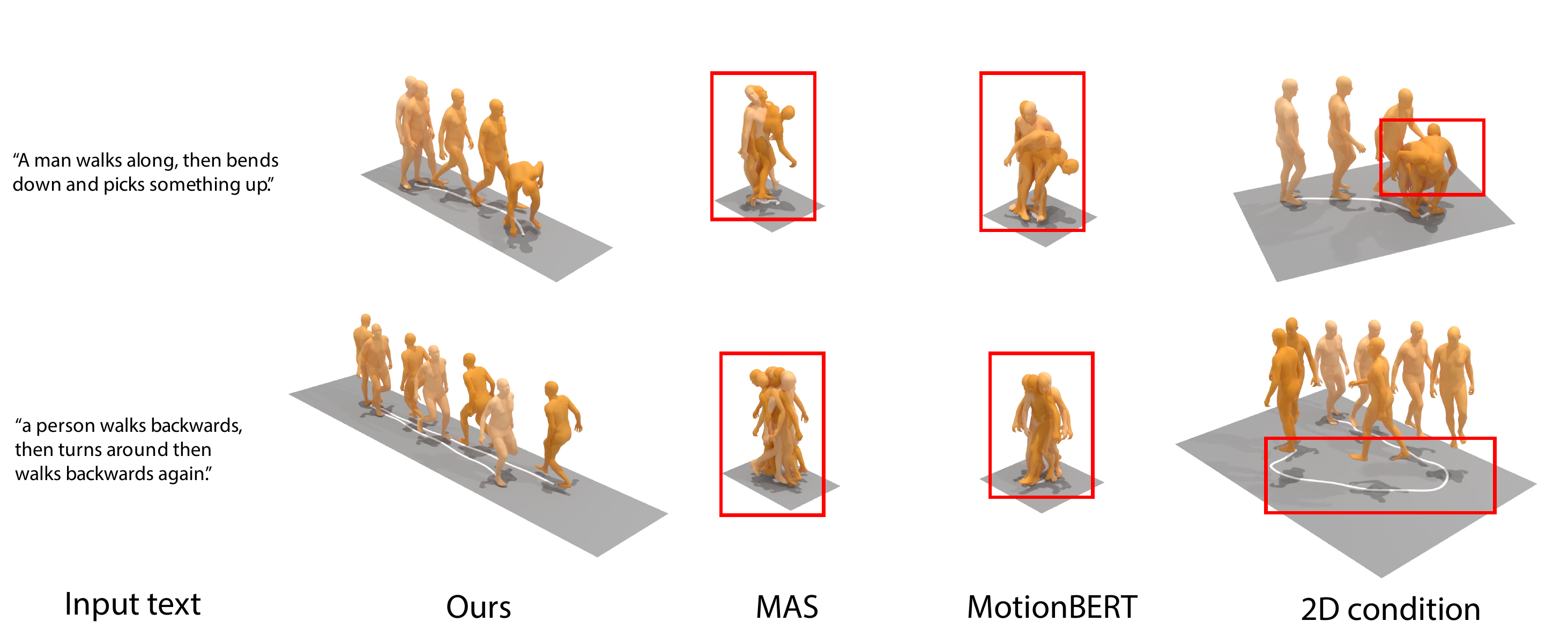}
    \vspace{0ex}
    \caption{\textbf{Qualitative results of different strategies using 2D data.}
    Baseline methods \cite{24cvpr_mas,23iccv_motionbert} fail to generate a motion with a global movement.
    The variant using 2D condition as input \cite{23iccv_zero1to3} may generate incorrect root movements, leading to the floating motion.
    The unnatural poses are highlighted in the red boxes.
    }
    \label{fig:use2d}
\end{figure*}

\subsection{Ablation study}
\begin{table}[tbp]
    \centering
    \resizebox{0.9\linewidth}{!}{
    \begin{tabular}{lccccc}
    \toprule
    Methods & R-Precision $\uparrow$&
    FID$\downarrow$ &
    MM Dist$\downarrow$&
    Diversity$\rightarrow$\\
    \midrule
    Real                                                & $0.797$ & $0.002$  & $2.974$ & $9.503$ \\
    \midrule
    Motion-X++ 3D \cite{24nips_motionx}                                & $0.596$ & $0.893$ & $4.188$ & \cellsecond $9.734$ \\
    \midrule    
    MotionBERT \cite{23iccv_motionbert}                 & $0.334$ & $23.292$ & $6.537$ & $5.441$ \\
    MAS \cite{24cvpr_mas}                               & $0.418$ & $11.893$ & $5.606$ & $6.413$ \\
    2D condition \cite{23iccv_zero1to3}                 & \cellsecond $0.668$ & \cellsecond $0.877$  & \cellsecond $3.790$ & $9.189$ \\
    \midrule
    Ours                                                & \cellfirst $\textbf{0.697}$ & \cellfirst $\textbf{0.321}$  & \cellfirst $\textbf{3.579}$ & \cellfirst $\textbf{9.286}$ \\
    \bottomrule
    \end{tabular}
    }
    \caption{
    \textbf{Ablation study on data utilization strategies.} 
    (1) Directly using 3D mocap annotations for motion generation results in less natural motion.
    (2) Our proposed strategy of lifting 2D motion to 3D achieves more effective results compared to baseline methods.
    }
    \label{tab:use2d}
\end{table}
\paragraph{Comparison on using 3D mocap annotation.}
While leveraging web video collections for motion generation, a natural approach involves employing monocular 3D motion capture as the motion source. 
Our baseline experiments using Motion-X \cite{24nips_motionx} 3D annotations revealed suboptimal performance compared to our proposed method, as shown in \Tabref{tab:use2d}.
This is primarily due to two factors: (1) inherent instability in reconstructed 3D motion quality, and (2) unrealistic motion priors introduced by 3D motion estimation from monocular videos.
In contrast, our method uses a mixed strategy that first uses 2D motion data, which is less noisy than the 3D annotation, and then uses high-quality motion data for 3D motion prior learning.

\paragraph{Evaluating strategies for lifting 2D motion to 3D.}
To evaluate the effectiveness of our approach, we compare it against alternative strategies for lifting 2D motion to 3D:
(1) \textbf{MotionBERT}: This method generates 2D motion with \twodnet{}, then directly infers 3D motion from the generated 2D data using MotionBERT~\cite{23iccv_motionbert}.
(2) \textbf{MAS}~\cite{24cvpr_mas}: MAS uses \twodnet{} to independently generate multi-view 2D motions. It enforces consistency by projecting inferred 3D motions back into 2D space.
(3) \textbf{2D condition}: Following Zero-1-to-3~\cite{23iccv_zero1to3}, this method employs another \mvnet{} trained specifically to generate multi-view outputs from the 2D motions produced by \twodnet{}.

As shown in \Tabref{tab:use2d}, our proposed approach achieves superior performance compared to these baselines, highlighting its effectiveness. High FID scores for MotionBERT and MAS are due to their inability to capture global movements. Our approach not only addresses this limitation but also ensures multi-view consistency more effectively than MAS, generating coherent multi-view outcomes. We also find the 2D condition method less effective, as it fails to correct artifacts produced by \twodnet{}.
Qualitative results presented in \Figref{fig:use2d} confirm these findings. MAS and MotionBERT exhibit limited global movement, resulting in unrealistic motions, especially for walking scenarios. The 2D condition method also displays noticeable artifacts due to inaccuracies in initial 2D generation.

\begin{table}[tbp]
    \centering
    \resizebox{0.85\linewidth}{!}{
        \begin{tabular}{lccccc}
    \toprule
    Methods & R-Precision $\uparrow$&
    FID$\downarrow$ &
    MM Dist$\downarrow$&
    Diversity$\rightarrow$\\
    \midrule

    Real                                                & $0.797$ & $0.002$ & $2.974$ & $9.503$ \\
    \midrule
    w/o pretrain                                        & $0.545$ & $4.950$ & $4.717$ & $7.360$ \\
    w/ CB                                               & $0.679$ & $0.642$ & $3.723$ & \cellfirst $\textbf{9.522}$ \\
    View=3                                              & $0.689$ & $0.654$  & $3.607$ & $8.946$ \\
    View=5                                              & \cellsecond $0.691$ & $\cellsecond 0.593$  & \cellsecond $3.598$ & $9.042$ \\
    \midrule
    Ours                                                & \cellfirst $\textbf{0.697}$ & \cellfirst $\textbf{0.321}$  & \cellfirst $\textbf{3.579}$ & \cellsecond $9.286$ \\
    \bottomrule
    \end{tabular}
    }
    \caption{
    \textbf{Ablation study on model designs.}
    }
    \label{tab:ablation}
\end{table}
\paragraph{Model designs.}
We evaluate four variants to validate the model design.
(1) w/o pre-training: \mvnet{} is trained from scratch without using \twodnet{} as a pretrained model.
(2) w/ CB: we incorporate the consistency block from \cite{24cvpr_mas} into \mvnet{}’s diffusion process, performing triangulation to convert multi-view outputs into 3D motion at each diffusion step, which is then projected back to each view.
(3) View=3: \mvnet{} is trained with only $3$ views.
(4) View=5: \mvnet{} is trained with $5$ views.
The results are shown in \Tabref{tab:ablation}.
We could observe that without pretraining on 2D motion, the performance of the model drops significantly, indicated by the much higher FID.
We find that the consistency block does not improve performance in our setting, possibly because \mvnet{} already captures view consistency effectively.
Additionally, at early diffusion steps, imprecise predictions may cause the consistency block to introduce extra noise. 
Empirically, using $4$ views in \mvnet{} yields the best performance.

\subsection{Limitation and future work}
While our model shows promising results, several limitations remain.
2D motion extracted from videos may contain noise and jitter, potentially affecting the quality of the generated 3D motion.
Although we leverage public 2D videos from existing datasets, which offer a greater scale than 3D motion datasets, this scale is still limited compared to video generation \cite{23arxiv_svd}, suggesting that further scale-up could enhance performance.
Our architecture, similar to MDM \cite{22iclr_mdm}, could benefit from exploring other neural network designs, such as GPT \cite{23cvpr_t2mgpt}.
Extending our method to hand motion, which is challenging to capture in MoCap systems \cite{21_easymocap}, and incorporating object interactions \cite{24cvpr_mas} could also enhance realism and diversity in generated motions.
\section{Conclusion}
\label{sec:conclusion}
We introduce \shortname{}, a novel pipeline to generate 3D motion from text input.
\shortname{} employs a root decoupled multi-view 2D motion representation to bridge the gap between 2D and 3D motion.
Leveraging this representation, our method first trains a 2D local motion generator on a large dataset of 2D motion extracted from videos. 
Subsequently, it finetunes the generator with 3D data to create a multi-view model capable of predicting view-consistent 2D motion and root velocities.
Experimental results demonstrate that our pipeline obtains better performance and broadens the range of motion types it can generate.
Our work not only demonstrates the potential of integrating 2D motion data into 3D motion synthesis but also opens up new possibilities for leveraging large-scale 2D motion datasets to advance human motion generation.

\paragraph{Acknowledgements.}\small{This work was partially supported by the following grants: National Key R \& D Program of China (No. 2024YFB2809102), Zhejiang Provincial Natural Science Foundation of China (No. LR25F020003), Deep Glint, Information Technology Center and State Key Lab of CAD \& CG, Zhejiang University.}
\end{document}